\documentclass[lettersize,journal]{IEEEtran}
\usepackage{amsmath,amsfonts}
\usepackage{algorithmic}
\usepackage{algorithm}
\usepackage{array}
\usepackage[caption=false,font=normalsize,labelfont=sf,textfont=sf]{subfig}
\usepackage{textcomp}
\usepackage{stfloats}
\usepackage{url}
\usepackage{verbatim}
\usepackage{graphicx}
\usepackage{color}
\usepackage{cite}
\usepackage{booktabs}
\usepackage{framed,multirow}

\usepackage{booktabs}
\usepackage{colortbl}

\usepackage[T1]{fontenc}

\usepackage{amsthm,amsmath,amssymb}
\usepackage{mathrsfs}

\newcommand{\he}[1]{\textcolor{red}{{}#1}}
\begin{document}

\title{Learning Progressive Adaptation for Multi-Modal Tracking}

\author{He Wang, Tianyang Xu,~\IEEEmembership{Member,~IEEE}, Zhangyong Tang, Xiao-Jun Wu,  Josef Kittler, ~\IEEEmembership{Life Member,~IEEE}
\thanks{
This work is supported in part by the National Natural Science Foundation of China (Grant NO. 62020106012, 62576152, 62332008), the Basic Research Program of Jiangsu (BK20250104), the Fundamental Research Funds for the Central Universities (JUSRP202504007), the 111 Project of Ministry ofEducation of China (Grant No.B12018), and the UK EPSRC (EP/N007743/1.MURI/EPSRC/DSTL, EP/RO18456/1).

He Wang, Tianyang Xu, Zhangyong Tang, and Xiao-Jun Wu (Corresponding author) are with the School of Artificial Intelligence and Computer Science, Jiangnan University, Wuxi 214122, China (e-mail: 7243115005@stu.jiangnan.edu.cn; tianyang.xu@jiangnan.edu.cn; zhangyong\_tang\_jnu@163.com; 
wu\_xiaojun@jiangnan.edu.cn).

Josef Kittler is with the Centre for Vision, Speech and Signal
Processing, University of Surrey, GU2 7XH Guildford, U.K. (e-mail: j.kittler@surrey.ac.uk).}}


\markboth{Journal of \LaTeX\ Class Files,~Vol.~14, No.~8, August~2021}%
{Shell \MakeLowercase{\textit{et al.}}: A Sample Article Using IEEEtran.cls for IEEE Journals}


\maketitle

\begin{abstract}
Due to the limited availability of paired multi-modal data, multi-modal trackers are typically built by adopting pre-trained RGB models with parameter-efficient fine-tuning modules. However, these fine-tuning methods overlook advanced adaptations for applying RGB pre-trained models and fail to modulate a single specific modality, cross-modal interactions, and the prediction head. To address the issues, we propose to perform Progressive Adaptation for Multi-Modal Tracking (PATrack). This innovative approach incorporates modality-dependent, modality-entangled, and task-level adapters, effectively bridging the gap in adapting RGB pre-trained networks to multi-modal data through a progressive strategy. Specifically, modality-specific information is enhanced through the modality-dependent adapter, decomposing the high- and low-frequency components, which ensures a more robust feature representation within each modality. The inter-modal interactions are introduced in the modality-entangled adapter, which implements a cross-attention operation guided by inter-modal shared information, ensuring the reliability of features conveyed between modalities. Additionally, recognising that the strong inductive bias of the prediction head does not adapt to the fused information, a task-level adapter specific to the prediction head is introduced. In summary, our design integrates intra-modal, inter-modal, and task-level adapters into a unified framework. Extensive experiments on RGB+Thermal, RGB+Depth, and RGB+Event tracking tasks demonstrate that our method shows impressive performance against state-of-the-art methods. Code is available at https://github.com/ouha1998/Learning-Progressive-Adaptation-for-Multi-Modal-Tracking. 
\end{abstract}


\section{Introduction}
\IEEEPARstart{T}{he} advancement of RGB single object tracking (SOT) has been significantly driven by the ability of RGB images to provide rich textural information about objects in typical settings, supported by the economic feasibility of the necessary sensor technology.
Recent RGB SOT trackers \cite{ostrack, STARK, siamfc, dimp, xu2025less, transt, odtrack, ProbabilisticRegressionforVisualTracking,SiamCorners} have demonstrated strong tracking capabilities.
However, when confronted with challenging visual environments like low illumination, occlusions, or fast-moving objects, the capability of RGB images to provide sufficient target information is limited.
Therefore, enhancing model robustness through supplementary modalities that can compensate for degraded RGB image quality is essential for accurate object localization.
Depth, thermal, and event modalities \cite{rgbd1k,melt,tenet} have all demonstrated their effectiveness in multi-modal information fusion.  
Recently, several approaches 
\cite{GMMT, onetracker,zhu2024unimod1k} 
have been developed for {multi-modal object tracking}, incorporating various sensor combinations RGB+X, such as RGB+Thermal, RGB+Event, or RGB+Depth.
{To explore the supportive potential of existing RGB trackers in RGB+X scenarios, our work only focuses on inputs comprising RGB+Thermal, RGB+Depth, or RGB+Event modalities.}
These approaches of combining multi-modal image information have garnered widespread attention in real-world applications, especially in safety-critical scenarios like autonomous driving \cite{unmanneddriving}.
Moreover, they have seen significant advancements in other domains, including semantic segmentation \cite{zhuang2021perception} and object detection \cite{jiang2022prototypical}.

Most previous studies \cite{tbsi, rgbd1k, det} utilize a global fine-tuning strategy, adapting RGB-based trackers  \cite{zhu2021robust} into dual-branch symmetric multi-modal trackers.
While this full fine-tuning approach enables cross-modal learning, it increases the burden of storing trainable parameters and raises computational costs.
It is essential to study {multi-modal object tracking} using approaches that avoid full fine-tuning, as this reduces computational overhead and enhances models' adaptability to diverse modalities without compromising efficiency.

\begin{figure}[t]
\centering
\includegraphics[width=0.85\linewidth]{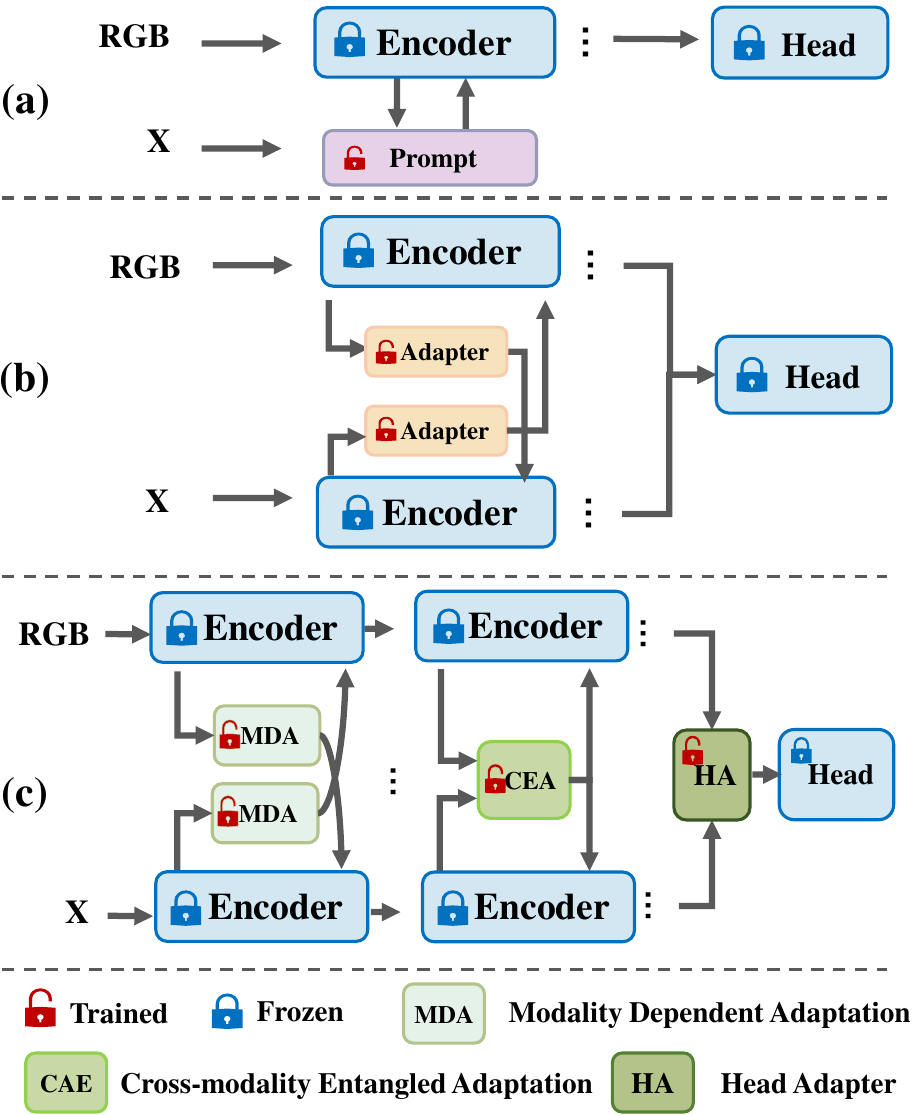}
\caption{
{A comparison between the original fusion mechanisms and our proposed fusion mechanism: (a) The asymmetric structure reliant on the dominant modality, employing the prompt fine-tuning paradigm; (b) The symmetric structure that considers the dominant and auxiliary modalities as equal, emphasizing the complementary information across modalities; and (c) our proposed method (PATrack) incorporates progressive adaptation learning at three levels: intra-modality, inter-modality, and task level. X represents thermal, event, or depth input. }
}
\label{fig:1}
\end{figure}

{Subsequently, a partial fine-tuning paradigm \cite{protrack, C2C, Madapter,vipt,sdstrack} appears, addressing the previously mentioned challenges by freezing the foundation network and introducing only a minimal set of trainable parameters.}
While deploying two dual camera sensors offers increased sensory data, balancing the substantial spatial differences between RGB and other modalities constitutes a significant challenge for researchers in multi-modal object tracking.
Generally, multi-modal trackers utilizing adapter-based or prompt-based fine-tuning mechanisms are categorized into symmetric and asymmetric structures.
The approach where one modality serves as the primary information and another modality as auxiliary information is termed an asymmetric tracker (see Fig. \ref{fig:1} (a)), such as \cite{protrack,vipt}. 
However, the tracking performance of the asymmetric structure network can be severely affected when images of the dominant modality are degraded or lose effective information.
Conversely, symmetric trackers aim to integrate complementary multi-modal information, minimizing dependency on a dominant modality.
Recent symmetric methodologies \cite{bat} (Fig. \ref{fig:1} (b)) utilize a bi-directional adapter to facilitate specific feature extraction. 
Nevertheless, because this specific modality adapter does not facilitate cross-modality fusion, it overlooks inter-modality correlation information, increasing the potential for noise.

To address these issues, we propose a progressive adapter-tuning paradigm that integrates joint training at the intra-modality, inter-modality, and task level for RGB+X tracking, as illustrated in Fig.~\ref{fig:1} (c).
Our method aims to enhance the utilization of modality-complementary information and the potential for shared representations between RGB and X data.
Leveraging modality-complementary information can compensate for the deficiencies in specific modalities.
{By introducing Modality-Dependent Adaptation (MDA), a modality-specific adapter that incorporates both high- and low-frequency information, we aim to improve the modality-complementary representation.}
{To extract reliable target information from the most discriminative features across modalities, we also introduce Cross-Modality Entangled Adaptation (CEA), which leverages cross-modal correlations to complement the MDA module.}
The combined application of CEA and MDA significantly enhances the network robustness and accuracy, as shown in Fig. \ref{fig:1.1}.
{Furthermore, the strong inductive bias \cite{rrn,Relationalinductivebiases} present in the prediction head hinders effective adaptation to multi-modal scenarios.} 
Therefore, we introduce Head Adaptation to address the issue, a task-level adapter that modulates the prediction head, effectively filtering out noise and background information. 

Our method combines Modality-Dependent Adaptation, Cross-Modality Entangled Adaptation, and Head Adaptation to construct a progressively adapted RGB+X tracker, further enhancing the adaptability of RGB pre-trained models for multi-modal data.
In summary, our contributions can be summarised as follows:
\begin{itemize}
	\item
    {A progressive adaptation strategy is proposed to refine multi-modal features, thereby enhancing the adaptability of the foundation model pre-trained in the RGB domain and providing robust feature enhancement for multi-modal tracking tasks.}
    \item 
    The novel Modality-Dependent Adaptation and Cross-Modality Entangled Adaptation are proposed to enhance modality-complementary information and facilitate cross-modal shared information. {We also develop a task-level adaptation for the prediction head to adapt further networks pre-trained on RGB data to multi-modal tracking tasks.}
	\item 
    Extensive experimental results demonstrate the superior performance of our method across the RGB-T, RGB-D, and RGB-E tracking fields.
\end{itemize}

\section{Related Work}
In this section, we provide an overview of the technologies closely related to our work, including {visual object tracking}, multi-modal tracking and the application of Parameter-efficient Fine-tuning (PEFT) \cite{Parameter-EfficientTransferLearningforNLP} techniques on multi-modal tracking tasks. 

\subsection{{Visual Object Tracking}} In recent years, substantial research efforts \cite{ostrack,STARK,ATOM,dimp,cheng2025fusionbooster} have focused on visual object tracking, which has found wide-ranging applications across various fields. 
{However, under extreme conditions, such as low lighting or fog, relying solely on RGB data greatly compromises tracking accuracy. }
{This limitation has driven the exploration of alternative sensing modalities, each offering unique advantages: thermal imaging \cite{dfat,apfnet,Lu_Li_Yan_Tang_Luo_2021,sttrack,taat,MaCNet,mv-rgbt} provides robustness in low-light conditions but suffers from thermal crossover and transparency challenges; event cameras \cite{tenet} offer exceptional temporal resolution for high-speed motion capture while lacking detailed texture information; and depth sensing effectively handles occlusion scenarios but cannot provide color cues \cite{zhang2021object,zhu2024pr}.
Since each thermal, event, depth modality provides unique complementary advantages to RGB data, this has led to the natural emergence of multi-modal tracking frameworks.
}

\subsection{Multi-modal Tracking}
{Multi-modal learning \cite{MFGNet,cbpnet} is a rapidly advancing research area within machine learning and artificial intelligence, showing significant promise across diverse applications such as autonomous driving, mobile robotics, video surveillance, and human-robot interaction.
{An increasing number of research is now exploring multi-modal tracking to address the limitations of relying solely on RGB data.}
For instance, RGB-T tracking \cite{GMMT} overcomes the sensitivity of RGB images to illumination by incorporating thermal data.
{RGB-D tracking \cite{rgbd1k} enhances the visibility of occluded objects by incorporating depth information, while RGB-E tracking \cite{visevent}, with its low-latency motion capture capabilities (1 \textmu s) provided by event sensors, improves tracking performance in low-light and high-speed scenarios by utilizing event flows.}
Also, various datasets are constructed to advance the development of multi-modal tracking. 
{Wang et al. propose high-resolution datasets, including CRSOT \cite{crsot} with 1,030 unaligned RGB-E image pairs and EventVOT \cite{Eventstream-basedvisualobjecttracking} with 1,141 aligned RGB-E image pairs, respectively.}

{Effectively fusing diverse data is crucial for achieving accurate performance in multi-modal tracking.}
Early CNN-based models, like MFGNet \cite{MFGNet}, aim to improve modality interaction but are limited by the inherent constraints of CNN architectures.
{With the rapid advancement of deep neural networks, recent studies \cite{protrack, Luo_Dong_Guo_Yu_2023} turn to Transformer-based architectures \cite{transformer} for feature extraction, demonstrating substantial improvements in both accuracy and robustness. 
The flexibility of the Transformer architecture allows for more complex fusion strategies, enabling better integration of multi-modal data \cite{GMMT,sdstrack,RGBTSalientObjectDetection}. 
Despite these advances, a major hurdle remains the scarcity of annotated multi-modal datasets, which limits the ability to train robust models. 
As a result, many current approaches {\cite{mat,emtrack}}, \cite{protrack,vipt,bat} apply fine-tuning pre-trained networks based on RGB data for multi-modal tasks. 
However, these methods lack exploration of the adaptability of the RGB-based pre-trained model for multi-modal networks.
Our approach employs an adapter fine-tuning paradigm to investigate the adaptability of pre-trained networks to downstream tasks at a more granular level, effectively aligning multi-modal data with spaces pre-trained on RGB data.}

\subsection{Parameter-efficient Fine-tuning in Multi-Modal Tracking}
{Recent advances in large models \cite{bert} demonstrate remarkable generalization capabilities across various domains. 
These models typically employ complex architectures trained on extensive datasets, achieving great performance in numerous vision tasks. However, when applied to multi-modal object tracking, a fundamental challenge emerges.}

{The primary challenge lies in modality adaptation, particularly when transferring knowledge from RGB-based models to other modalities. 
Thermal infrared tracking \cite{liu2020lsotb} faces significant obstacles due to the modality's inherent characteristics, particularly its sensitivity to temperature variations and lack of detailed texture information compared to RGB data.
1While domain adaptation approaches \cite{li2025efficient} can mitigate these issues, they remain constrained by the fundamental differences in data distribution between modalities.}

{Recent works explore parameter-efficient fine-tuning to address modality adaptation. A common approach employs asymmetric architectures that treat RGB as the dominant modality while incorporating auxiliary data through prompt-tuning \cite{protrack,vipt}. 
However, these methods often struggle to fully leverage complementary modal information, particularly in capturing critical inter-modal relationships. 
The limitations become especially apparent in complex scenarios where robust multi-modal integration is crucial.}

{Although subsequent symmetric architectures \cite{sdstrack,bat} attempt to treat RGB and auxiliary modalities as equally important, thereby advancing effective modality fusion, these approaches still exhibit two fundamental limitations. First, they primarily focus on modality-specific information extraction while largely neglecting the critical inter-modal shared features that could enhance tracking robustness. Second, existing methods fail to address the inherent challenge of adapting RGB-based pre-trained parameters to multi-modal tasks, particularly due to the significant distribution discrepancies between different modalities.
}
{To address these issues, we propose a progressive adaptation framework for multi-modal tracking. This framework incorporates modality-dependent, modality-entangled, and task-level adapters, providing a comprehensive solution to the aforementioned challenges.}


\begin{figure*}[t]
\centering
\includegraphics[width=1\textwidth]{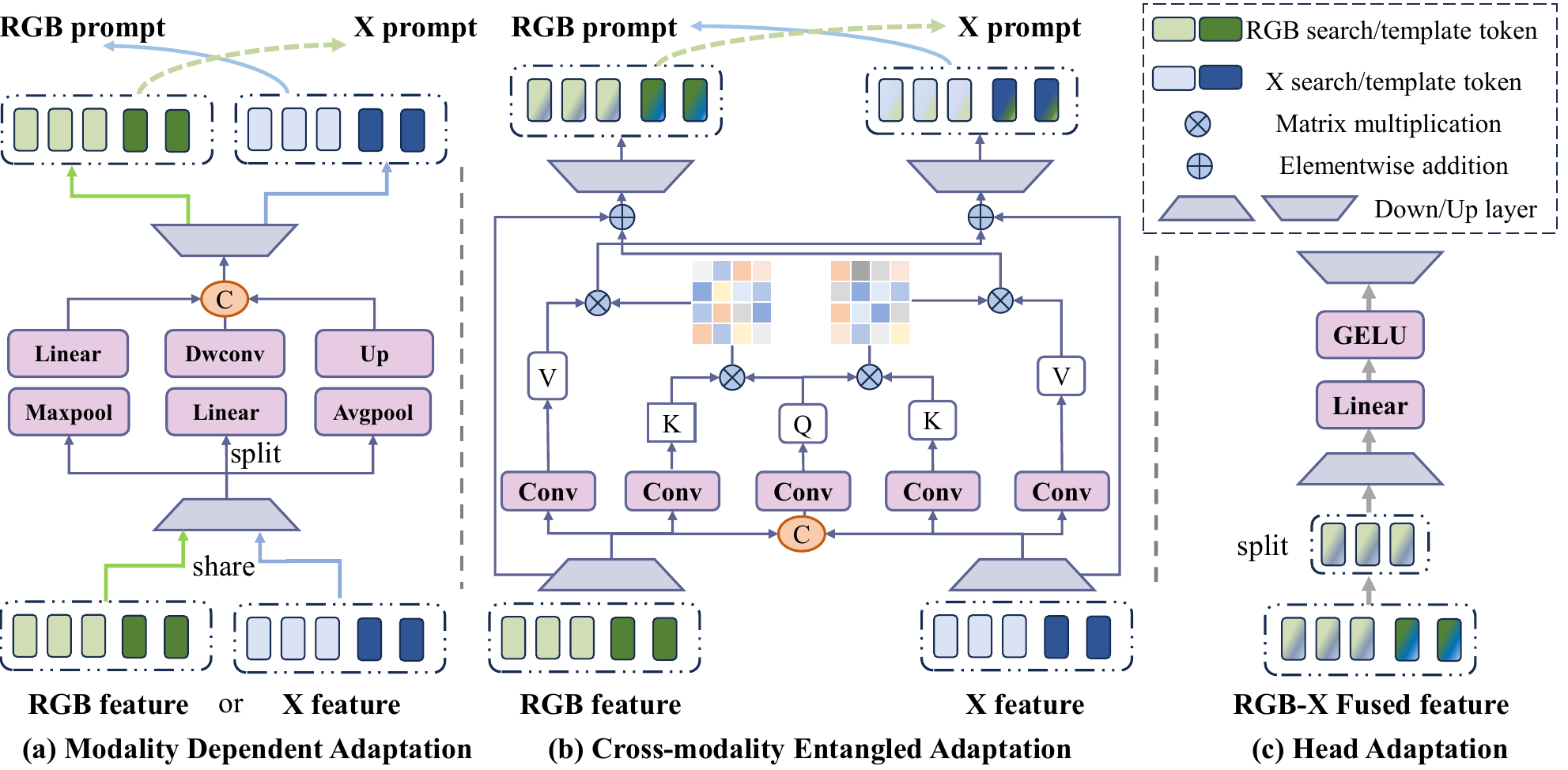}
\caption{Illustration of the proposed PATrack. In the Modality Dependent Adaptation (MDA) module, only one modality can be sent to this architecture. The term “share” indicates that the RGB and X branches share parameters.}
\vspace{-3mm}
\label{fig:2}
\end{figure*}

\section{Methodology}
In this section, we initially provide an overview of our foundation network, specifically the RGB Base Model developed by OSTrack \cite{ostrack}. 
We extend this model by integrating an additional modality for RGB-X tracking tasks, where X can represent the thermal, depth, or event modality.
Each task is trained independently using the dataset corresponding to its respective modality.
Following this, we introduce our Modality-Dependent Adaptation and Cross-Modality Entangled Adaptation mechanisms in detail. We then describe the Head Adaptation at the task level, a process designed to further reduce the disparity between the pre-trained model and the multi-modal task, effectively filtering out noise during feature extraction.
The overall pipeline is illustrated in Fig. \ref{fig:1} (c), with details of the network components presented in Fig. \ref{fig:2}.

\subsection{RGB Base Model}
In the field of visual object tracking using only the RGB modality, the object tracker $Tracker$ is given the initial bounding box position $B_o$ of the target from the template frame $T_{\text{RGB}} \in \mathbb{R}^{H_T \times W_T \times 3}$. 
The $Tracker$ is then trained to locate this target in subsequent search frames $S_{\text{RGB}} \in \mathbb{R}^{H_S \times W_S \times 3}$.
Typically, the $Tracker$ comprises a feature extraction function $F$ and a box head Head.

The RGB template frame $T_{\text{RGB}}$ and the RGB search frame $S_{\text{RGB}}$ are together processed by a foundation model based on a Transformer architecture. 
And then they are converted into tokens using the patch embedding layer and they combine with a position embedding. 
After concatenating template tokens and search tokens together, multi-modal features are fed into $N$-layer Transformer encoder blocks to jointly extract features.
The search region features are disassembled from output tokens along the channel dimension and then fed into the prediction Head architecture to generate the target tracking result.
The final location of the tracking target ${B}_{box}$ is predicted as follows:

\begin{equation}
   \centering \label{equ:1}
{{B}_{box} = \mathrm{Head}(F({S}_{\text{RGB}},{T}_{\text{RGB}},{B}_{o})),}
\end{equation}

where $F$ employs the powerful OSTrack \cite{ostrack} backbone trained on RGB data to extract features, whose architecture includes multihead self-attention (MSA), layer normalization (LN), feedforward neural network (FFN), and residual connections.

\subsection{Multi-modal Tracking}
We propose Progressive Multi-modal Adaptation for Downstream Multi-modal Tracking tasks (PATrack), a framework that effectively transfers RGB-based pre-trained networks to downstream multi-modal tasks with minimal training parameters and computational load by leveraging adapter learning.
Specifically, PATrack incorporates two mutually complementary components: Modality-Dependent Adaptation (MDA) and Cross-modality Entangled Adaptation (CEA). 
These modules jointly extract and fuse multi-modal features, ensuring efficient and robust feature integration. 
Furthermore, recognizing the limited adaptability of the RGB-based pre-trained prediction head for processing multi-modal data and accurately predicting the target bounding box, we introduce a head-specific adapter structure to address this limitation effectively.

Multi-modal object tracking expands the RGB-based framework by integrating multiple modalities through the introduction of an additional modality stream. This multi-modal approach collaboratively contributes to the final prediction of the target's bounding box, denoted as ${B}_{box}$. 
{Given that the RGB modality and the auxiliary modality (X, e.g., Thermal, Depth, or Event) are generally spatially aligned and temporally synchronized, the target's position in the first frame is identical across modalities and is denoted as $B_{o}$.}
Typically, the $Tracker$ network comprises a feature extraction function $F$ and a box head Head.
During tracking, a pair of search and template frames consisting of search frames ${S}_{\text{{RGB}}}$, ${S}_{\text{X}}$ $ \in \mathbb{R}^{{H}_{S} \times {W}_{S} \times 3}$ and template frames \({T}_\text{{RGB}}\), \({T}_\text{{X}} \in \mathbb{R}^{{H}_{{T}} \times {W}_{{T}} \times 3}\) are sent into the foundation network.
The final location ${B}_{box}$ of the target is predicted as follows:

\begin{equation}
    \centering \label{equ:2}
{B}_{box} = \text{{Head}}(F({T}_\text{{RGB}},{T}_\text{{X}},{S}_\text{{RGB}},{S}_\text{{X}},{B}_{0}))\he{,}
\end{equation}

As illustrated in {Fig. \ref{fig:1} (c)}, PATrack adopts a dual-branch architecture that incorporates both the RGB modality and an auxiliary modality (Thermal, Depth, or Event). 
This design effectively addresses the limitations posed by the auxiliary modality's dependence on pre-trained RGB data by enabling the auxiliary branch to leverage the strengths of the RGB-based pre-trained network.
Furthermore, the parameters between the RGB and auxiliary modality (X) branches are shared, significantly enhancing the model computational efficiency and overall effectiveness.

The PATrack framework begins by passing the input data, ${S}_\text{{RGB}}$, ${S}_\text{{X}}$, ${T}_\text{{RGB}}$, and ${T}_\text{{X}}$, through the patch embedding layer.
This step generates initial RGB tokens ${H}_\text{{RGB}}^{0}$ and auxiliary modality tokens ${H}_\text{{X}}^{0}$.
Subsequently, tokens from both modalities are independently processed in their respective designated branches.
During the layer-wise integration process, the $l$-th encoder layer receives RGB and X tokens generated by the previous $(l-1)$-th layer, as shown below:


\begin{equation}
\centering \label{equ:3}
({H}^{l}_\text{{RGB}}, {H}^{l}_\text{{X}}) = {{F}_{A}}^{l}({H}^{l-1}_\text{{RGB}}, {H}^{l-1}_\text{{X}}), l = 1, 2, ..., N\he{,}
\end{equation}

{Here, ${F}_{A}^{l}$ represents a dual-stream encoder layer implemented within our proposed adapter modules including Modality-Dependent Adaptation or Cross-modality Entangled Adaptation framework.}
This advanced layer is engineered to parallelize two encoder streams, facilitating the efficient capture and integration of complementary multi-modal information and shared data features.


As illustrated in {Fig. \ref{fig:1} (c)}, parallel adapter modules, including Modality-Dependent Adaptation or Cross-modality Entangled Adaptation, are inserted into the multi-head self-attention ($MSA$) and the multilayer perceptron ($MLP$) between Encoder blocks. 

The processing of different layers, including the MDA and CEA modules, is distinctly designed to optimize their respective functionalities.
We begin by describing the layers that incorporate the MDA module. 
Using the auxiliary modality branch as an example, the computation process is defined as follows:


\begin{equation}
\centering \label{equ:4}
{H}^{l'}_\text{{X}} = {H}^{l-1}_\text{{X}}+ {MSA}(LN({H}^{l-1}_\text{{X}}))+{Ada}^{l-1}({H}^{l-1}_\text{{RGB}})\he{,}
\end{equation}

{Here, $Ada$ represents the adaptation module, which can be configured for either the MDA or CEA, depending on the layer's requirements.
The same presentation of $Ada$ applies in subsequent contexts.} The converse operation yields ${H}^{l'}_\text{{RGB}}$.

Subsequently, ${{H}^{l'}_\text{{X}}}$ is fed into the $MLP$ layer, and combine with adapter feature ${Ada}^{l-1}({H}^{l'}_\text{{RGB}})$ to obtain the output ${{H}^{l}_\text{{X}}}$ in the $l$ -th layer. 
\begin{equation}
\centering \label{equ:5}
{H}^{l}_\text{{X}} = {H}^{l'}_\text{{X}} + {MLP}(LN({H}^{l'}_\text{{X}}))+{Ada}^{l-1}({H}^{l'}_\text{{RGB}})\he{,}
\end{equation}

where $l \in \{1, 2, 3, 5, 6, 8, 9, 11, 12\}$.

The RGB branch acquires ${H}^{l}_{RGB}$ in the opposite way.

The processes including the CEA module are as follows:
\begin{equation}
\centering \label{equ:4.1}
{H}^{l'}_\text{{X}} = {H}^{l-1}_\text{{X}}+ {MSA}(LN({H}^{l-1}_\text{{X}}))+{Ada}^{l-1}({H}^{l-1}_\text{{X}})\he{,}
\end{equation}

\begin{equation}
\centering \label{equ:5.1}
{H}^{l}_\text{{X}} = {H}^{l'}_\text{{X}} + {MLP}(LN({H}^{l'}_\text{{X}}))+{Ada}^{l-1}({H}^{l'}_\text{{X}})\he{,}
\end{equation}

where $l \in \{4,7,10\}$.

\subsection{Modality-Dependent Adaptation}

As shown in Fig. \ref{fig:2} (a), the Modality-Dependent Adaptation (MDA) module follows a bottleneck design.
{The MDA module only operates a single modality at a time, either the RGB modality or the X modality. 
The processing operations for the two modalities are identical, but do not share parameters.
Initially, features are passed through a downsampling layer to reduce their dimensionality from $C$ to $C'$. 
In our case, $C$ is set to 768 and $C'$ is set to 8.
}

{Taking the X modality as an example, the low-dimensional feature map, $H_\text{X}$, is split along the channel dimension into ${H}_{X-h}\in {R}_n\times{C}_{h}$, and ${H}_{X-l}\in {R}_n\times{C}_{l}$. ${C}_{h}$ = ${C}_{l}$ = $C'/2$ = 4.
Then ${H}_{X-h}$ and ${H}_{X-l}$ are sent to extract high-frequency and low-frequency information, respectively.}

Inspired by \cite{Inceptiontransformer}, we employ a combination of operations to extract frequency-specific features. The max-pooling operation ($Maxpool$) and depthwise separable convolution layer ($DwConv$) are used to extract high-frequency information, while the average pooling operation ($AVG$) is utilized to capture low-frequency information.


{${H}_{X-h}$ is also split into two halves, ${H}_{X-hm}$ and ${H}_{X-hd}$, along the channel dimension, where the channel transformation ${C}_{hm}={C}_{hd}={C}_{h}/2$. Subsequently, ${H}_{X-hm}$ passes through a max-pooling operation ($Maxpool$) and a linear layer ($FC1$), as described below:}

\begin{equation}
\centering \label{equ:6}
{Y}_{X-hm} = {FC1}(Maxpool({H}_{X-hm}))\he{,}
\end{equation}
{where the $Maxpool$ applies a two-dimensional max pooling layer with a kernel size of 3, a stride of 1, and padding of 1. Following the $Maxpool$, the feature dimensions remain unchanged. We employ a linear layer $FC1$ implemented as a 1×1 convolution, doubling the original channel dimension. So finally the channel dimension of ${Y}_{X-hm}$ is $C'/2$.}

${H}_{X-hd}$ is processed through a linear layer ($FC2$) and a depth-wise convolutional layer ($DWConv$) simultaneously.
The $DWConv$ utilizes a kernel size of 3, with a stride of 1 and padding of 1, ensuring the number of channels remains unchanged. {To address the absence of inter-channel interactions in $DWConv$, a fully connected layer $FC2$, implemented as a 1×1 convolution, is added before $DWConv$ to double the channel dimension:}

\begin{equation}
\centering \label{equ:7}
{Y}_{X-hd} = {DWConv}(FC2({H}_{X-hd}))\he{,}
\end{equation}
{where the channel dimension of ${Y}_{X-hd}$ is $C'/2$.}

Meanwhile, ${Y}_{X-l}$ is generated through an average pooling layer ($AVG$) and an up-sampling ($Up$) layer to capture low-frequency information:

\begin{equation}
\centering \label{equ:8}
{Y}_{X-l} = {Up}({AVG({H}_{X-l})}))\he{,}
\end{equation}
{where the channel dimension of ${Y}_{X-l}$ is $C'/2$. The spatial resolution is first reduced to half of the input feature 
${H}_{X-l}$'s resolution and then restored to its original dimensions through upsampling.}

{Subsequently, the high- and low-frequency components of a single modality are concatenated ($Concat$) and passed through an Up layer ($U$) to revert from $3C'/2$ to the original feature space dimension $C$:}

\begin{equation}
\centering \label{equ:9}
{Ada}({H}_{X}) = {U}({Concat}({Y}_{X-hm},{Y}_{X-hd},{Y}_{X-l})))\he{,}
\end{equation}

Conversely, the RGB branch undergoes a similar process to obtain ${Ada}({H}_{RGB})$.
Finally, the output of the MDA module from one modality is transferred to the other modality branch to achieve cross-modality fusion.

\subsection{Cross-modality Entangled Adaptation}
As previously discussed, {bi-directional} modality-dependent adaptation is designed to capitalize on the complementary information inherent across modalities. 
However, the importance of shared information within multi-modal features should be noticed in multi-modal object tracking.
Modality-shared information can help networks learn universal features across modalities, aiding in better generalization across diverse datasets and environments.
The interaction of modality-specific and complementary modality information across various hierarchical levels permits the network to adapt and synthesize information in a layered manner, thereby capturing the intrinsic attributes of the target more effectively. 
By introducing Cross-modality Entangled Adaptation (CEA) guided by multi-modal fusion representations, modality correlations are effectively exploited.
The joint training strategy of MDA and CEA enables the efficient utilization of both intra-modal and inter-modal features while preserving computational efficiency.

Fig. \ref{fig:2} (b) illustrates the details of the Cross-modality Entangled Adapter.
Initially, the downsampling layer $Down$ is conducted to perform dimension reduction on $H_\text{{RGB}}$ and $H_\text{{X}}$, resulting $\hat{H}_\text{{RGB}}$ and $\hat{H}_\text{{X}}$: 

\begin{equation}
\centering \label{equ:10}
{\hat{H}_\text{{RGB}}}, \hat{H}_\text{{X}} = {Down(H_\text{{RGB}}),Down(H_\text{{X}})}\he{,}
\end{equation}

We concatenate $\hat{H}_\text{{RGB}}$ and $\hat{H}_\text{{X}}$ to get fusion information.
Subsequently, a projection mechanism is implemented, encompassing convolutions, $Conv_{q}$, $Conv_{k}$, and $Conv_{v}$, with a 3 × 3 kernel size, to transform these multi-modal features into query, key and value components necessary for the cross-attention operation $CA$.
{These global features are subsequently integrated with the original features from an alternative branch. } 
The operations can be formulated as follows: 

\begin{equation}
\centering \label{equ:11}
{fus} = {Conv_{q}}({Concat(\hat{H}_\text{{RGB}},\hat{H}_\text{{X}})})\he{,}
\end{equation}

\begin{equation}
\centering \label{equ:12}
{H}^{'}_\text{{RGB}} = {H}_\text{{RGB}}+ {CA}(Conv_{v}(\hat{H}_\text{{X}}), Conv_{k}(\hat{H}_\text{{X}}), {fus})\he{,}
\end{equation}

\begin{equation}
\centering \label{equ:13}
{H}^{'}_\text{{X}} = {H}_\text{{X}}+ {CA}(Conv_{v}(\hat{H}_\text{{RGB}}),Conv_{k}(\hat{H}_\text{{RGB}}), {fus})\he{,}
\end{equation}

Upsampling layer $U$ scales ${{H}^{'}_\text{{RGB}}}$ and ${H}^{'}_\text{{X}}$ to the original channel dimension:

\begin{equation}
\centering \label{equ:14}
{Ada}({H}_\text{{RGB}}), {Ada}({H}_\text{{X}}) = U({H}^{'}_\text{{RGB}}), U({H}^{'}_\text{{X}})\he{,}
\end{equation}

\subsection{Head Adaptation}
In the field of visual object tracking, the box head, including a neural network structure, is used for detecting and localizing bounding boxes of targets in images.
The pre-trained foundation network incorporates a specialized box head that only processes RGB data, excelling in handling monocular images composed of colour information.
However, multi-modal tracking tasks, including RGB-D, RGB-E, and RGB-T, necessitate the handling of diverse data types that provide additional information beyond colour, such as depth, motion, or temperature.

\begin{table*}[]
\renewcommand{\arraystretch}{1.4}
\centering
\caption{A comparison with state-of-the-art methods on \textbf{LasHeR}, \textbf{DepthTrack}, and \textbf{VisEvent} benchmarks. Performance is denoted in \textcolor{red}{Red} for the best and in \textcolor{blue}{Blue} for the second-best, consistently throughout the table. The trackers marked with an '*' indicate the results obtained by retraining the network, consistently throughout the table.}\vspace{0.3mm} 
\label{tab1}\setlength{\tabcolsep}{1.8mm}\scalebox{1.1}{
\begin{tabular}{ccccccccccc}
\hline 
\multirow{2}{*}{Type} & \multirow{2}{*}{Method} & \multirow{2}{*}{Venue} & \multicolumn{3}{l}{LasHeR} & \multicolumn{3}{l}{DepthTrack} & \multicolumn{2}{l}{VisEvent} \\
 &  &  & \multicolumn{1}{l}{SR} & \multicolumn{1}{l}{PR} & \multicolumn{1}{l}{NPR} & \multicolumn{1}{l}{Pr} & \multicolumn{1}{l}{Re} & \multicolumn{1}{l}{F-score} & \multicolumn{1}{l}{SR} & \multicolumn{1}{l}{PR} \\ \hline \hline
\multirow{11}{*}{Separated} 
 & DeT \cite{det} & CVPR  2021 & - & - & - & 0.560 & 0.506 & 0.532 & - & - \\
 & OSTrack \cite{ostrack} & ECCV 2022 & 0.412 & 0.515 & - & 0.535 & 0.522 & 0.529 & 0.499 & 0.659 \\
 & APFNet \cite{apfnet} & AAAI 2022 & 0.362 & 0.500 & 0.439 & - & - & - & - & - \\
 & SPT \cite{rgbd1k}& AAAI 2023 & - & - & - & 0.527 & 0.549 & 0.538 & - & - \\ 
 & TBSI \cite{tbsi} & CVPR 2023 & 0.549 & 0.688 & 0.657 & - & - & - & - & - \\
 & {GMMT} \cite{GMMT}& AAAI 2024 & {{0.566}} &{{0.707}} & {{0.670}} & - & - & - & - & - \\
 & BAT* \cite{bat} & AAAI 2024 & 0.563 & 
 0.702 & - & 0.583 & 0.578 & 0.580 & 0.592 & 0.758 \\
 &{EventVOT} \cite{Eventstream-basedvisualobjecttracking} &CVPR 2024 &- &- &- &-&- &- &0.373 &0.546 \\
 & {CRSOT} \cite{crsot}&TMM 2025 &- &- &- &-&- &-&0.525 &0.741 \\
 & {MAT \cite{mat}} &Information Fusion 2025& 0.537 &0.670 & 0.630 &- &- &- &- &-\\
 &{PTrMA \cite{PTrMA}} &TIM 2025 &\textcolor{blue}{\textbf{0.568}} &\textcolor{blue}{\textbf{0.715}}&\textcolor{blue}{\textbf{0.675}}&- &- &- &- &-
 \\ \cline{1-11} 
\multirow{6}{*}{Unified} & ProTrack \cite{protrack} & ACMMM 2022 & 0.420 & 0.538 & - & 0.583 & 0.573 & 0.578 & 0.459 & 0.655 \\
 & ViPT \cite{vipt} & CVPR 2023 & 0.525 & 0.651 & - & 0.592 & 0.596 & 0.594 & 0.534 & 0.695 \\
 & {SDSTrack} \cite{sdstrack} & CVPR 2024 & 0.531 & 0.665 & - & \textcolor{red}{\textbf{0.619}} & \textcolor{red}{\textbf{0.609}} & \textcolor{red}{\textbf{0.614}} & \textcolor{blue}{\textbf{0.597}} & \textcolor{blue}{\textbf{0.767}} \\
& {EMTrack \cite{emtrack}} & TCSVT 2024 &0.533 &0.659&- &0.580 &0.585 &0.583 &0.584 &0.724 \\
& {M\textsuperscript{3}Track} \cite{M3Track} & SPL 2025 & 0.525 & 0.658 & 0.622 & 0.564 & 0.585 & 0.574 & 0.596 & 0.767\\
& PATrack & Our & \textcolor{red}{\textbf{0.578}} & \textcolor{red}{\textbf{0.718}} &\textcolor{red}{\textbf{0.683}}
 & \textcolor{blue}{\textbf{0.603}} & \textcolor{blue}{\textbf{0.597}} & \textcolor{blue}{\textbf{0.600}} & \textcolor{red}{\textbf{0.605}}& \textcolor{red}{\textbf{0.770}} \\ \hline
\end{tabular}}
\end{table*}

\begin{table*}[]
\renewcommand{\arraystretch}{1.3}
\centering
\caption{A comparison with state-of-the-art methods on \textbf{RGBT234} and \textbf{GTOT} benchmarks.}\vspace{0.3 mm} 
\label{tab2}\scalebox{1.15}{
\begin{tabular}{ccccccccccc}
\hline
\multicolumn{2}{c}{Method} & {MFGNet} 
& ProTrack & ViPT & TBSI  & SDSTrack & BAT &GMMT&\he{MAT}& PATrack \\
\multicolumn{2}{c}{Venue} & TMM
& ACMMM& CVPR & CVPR& CVPR & AAAI &AAAI&Information Fusion& Our \\
\multicolumn{2}{c}{Year} &2022& 2022 & 2023 &2023 &2024 &2024 &2024&2025  \\
\hline \hline
\multirow{2}{*}{RGBT234} & SR & 0.535 
& 0.599 & 0.617 & 0.637  & 0.625 &0.641&\textcolor{blue}{\textbf{0.647}}&0.631& \textcolor{red}{\textbf{0.651}} \\
 & PR & 0.783 
 & 0.795 & 0.835 & 0.871 & 0.848 & \textcolor{blue}{\textbf{0.868}} &\textcolor{red}{\textbf{0.879}}&0.847& \textcolor{red}{\textbf{0.879}} \\ \hline
\multirow{2}{*}{GTOT} & SR & 0.707
& - & - & -  & 0.760 & 0.763 &\textcolor{red}{\textbf{0.785}} &0.765& \textcolor{blue}{\textbf{0.783}} \\
 & PR & 0.889 
 & - & - & -  & 0.887 & 0.909 &\textcolor{red}{\textbf{0.936}} &0.910& \textcolor{blue}{\textbf{0.925}} \\ \hline
\end{tabular}}
\end{table*}

As illustrated in Fig. \ref{fig:2} (c), the Head Adaptation
helps transfer knowledge from pre-trained RGB data to multi-modal space, thereby accelerating the learning process and enhancing performance. 

Specifically, the multi-modal fusion data extracted includes template tokens ${T}_\text{{RGBX}}$ and search tokens ${S}_\text{{RGBX}}$. 
We need to filter out search tokens ${S}_\text{{RGBX}}$ and send them to the Head Adapter module. 
{This is a simple bottleneck design module, but it is very effective.}
Firstly, ${S}_\text{{RGBX}}$ undergoes a dimensionality reduction linear layer $Down$, followed by an activation layer $Act$ and an dimensionality enhancement linear layer $Up$:

\begin{equation}
\centering \label{equ:15}
{S}^{'}_\text{{RGBX}}= Up(Act(Down({S}_\text{{RGBX}})))\he{.}
\end{equation}

Then ${S}^{'}_\text{{RGBX}}$ is sent to the prediction head to obtain a predicted position of the target.
In summary, the primary role of the head adapter is to serve as a bridge, extending the capabilities of the box head, which is pre-trained on RGB data, to effectively process multi-modal data.
This extension enhances the performance of downstream visual object tracking tasks.

\section{Experiments}
\subsection{Experimental Setups}
In this section, we evaluate the advantages of our proposed method by comparing it against several of the latest multi-modal trackers. The selected trackers include BAT \cite{bat}, {MAT \cite{mat}}, {PTrMA \cite{PTrMA}}, APFNet \cite{apfnet}, OSTrack \cite{ostrack}, DeT \cite{det}, SPT \cite{rgbd1k}, TBSI \cite{tbsi}, and GMMT \cite{GMMT}, which are designed for separate multi-modal tracking tasks.
{Additionally, we compare against {M\textsuperscript{3}Track \cite{M3Track}}, {EMTrack \cite{emtrack}}, SDSTrack \cite{sdstrack}, ViPT \cite{vipt}, and ProTrack \cite{protrack}, which are unified multi-modal trackers capable of handling RGB-T, RGB-D, and RGB-E tracking.}

{Our proposed methodology is implemented in PyTorch and executed on a system with an Intel(R) Core(TM) i9-9980XE CPU and an NVIDIA 3090Ti GPU.}
The PATrack is trained on subsets of the LasHeR, VisEvent, and DepthTrack benchmarks for different downstream tasks.
The hyperparameters are finely tuned using the AdamW optimizer with a weight decay of  1e-4.
The backbone OSTrack \cite{ostrack} is trained by an initial learning rate of 4e-4, subject to an exponential decay with a decay ratio of 0.8.

The PATrack has accomplished the integration of a variety of downstream tracking tasks into a cohesive framework, including RGB-T tracking, RGB-E tracking, and RGB-D tracking.
Each task is trained independently, and during inference, the framework employs its respective set of network parameters specific to the training process.

\subsection{Datasets and Evaluation Metrics}
{Concerning RGB-D tracking, the tracker's efficacy is evaluated on Depthtrack \cite{det} datasets. }
{Moving on to RGB-T tracking, comparative evaluations are conducted on  LasHeR \cite{lasher}, RGBT234 \cite{RGBT234}, and GTOT \cite{GTOT} benchmarks.}
Finally, for RGB-E tracking, the experimental results are presented using the extensive VisEvent \cite{visevent} dataset. 
A comprehensive suite of metrics is employed to evaluate performance, including precision rate (PR), success rate (SR), normalized precision rate (NPR), recall (Re), and F-score. 
Across all tracking tasks, we uniformly apply three distinct adapter modules, ensuring consistency in experimental setups and facilitating fair comparisons.

\subsection{Comparison with State-of-the-art Approaches}

\textbf{LasHeR}. LasHeR \cite{lasher} dataset consists of 979 training video sequences and 245 testing video sequences for RGB-T tracking, with evaluations conducted using metrics such as Precision Rate (PR) and Success Rate (SR). Table \ref{tab1} presents the comparative analysis of our method against existing RGB-T trackers on the test dataset, highlighting the highest performance of our approach. 
{The PATrack establishes impressive results, outperforming EMTrack \cite{emtrack} and PTrMA \cite{PTrMA}, with improvements of 4.5\% and 1.0\% in SR, as well as  5.9\% and 0.3\% in PR, respectively.}

\begin{figure}[t]
\centering
\includegraphics[width=0.46\textwidth]{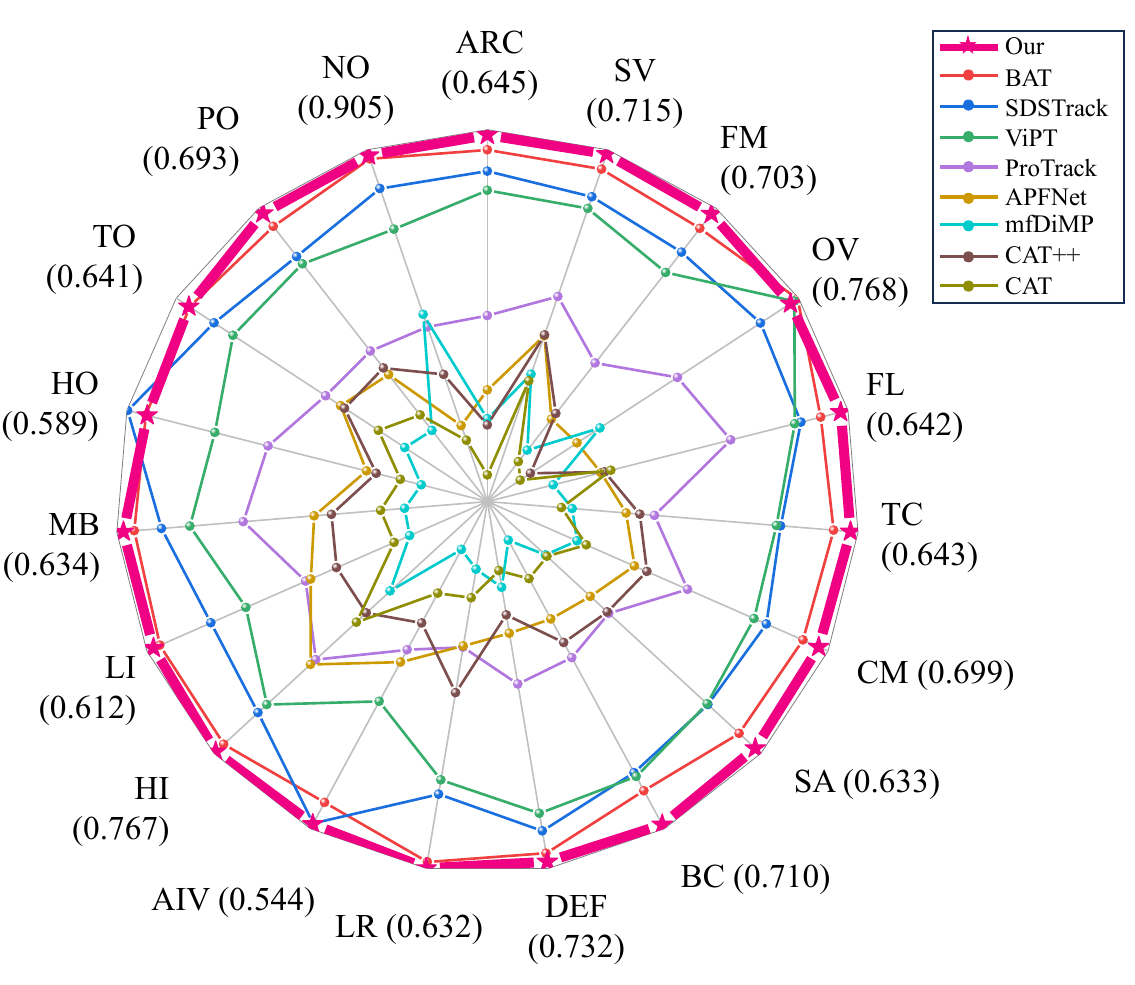} 
\caption{{Attribute-based Precision Rate on LasHeR dataset.}}
\label{fig3}
\end{figure}


\textbf{RGBT234.} {RGBT234 dataset \cite{RGBT234}, which integrates both RGB and thermal images, encompasses a total of 234 video sequences with nearly 116,700 frames.}
{Table~\ref{tab2} presents the performance of various trackers on RGBT234 dataset.}
Our proposed method achieves impressive results, with a Success Rate (SR) of 65.1\% and a Precision Rate (PR) of 87.9\%.
{These results surpass the second-best tracker, GMMT\cite{GMMT}, by 0.4\% in SR, and the third-best tracker, BAT\cite{bat}, by 1.0\% in SR and 1.1\% in PR.} 

\begin{table*}[]
\centering
\caption{The ablation experiment of our components on different RGB-X benchmarks.}
\vspace{1mm}
\label{tab7}
\renewcommand{\arraystretch}{1.1}
\scalebox{1.2}
{
\begin{tabular}{ccc|cc|ccc|cc|ccc}
\hline
\multirow{2}{*}{MDA} & \multirow{2}{*}{CEA} & \multirow{2}{*}{HA} & \multirow{2}{*}{\begin{tabular}[c]{@{}c@{}}Params\\ (M)\end{tabular}} & \multirow{2}{*}{\begin{tabular}[c]{@{}c@{}}Flops\\ (G)\end{tabular}} & \multicolumn{3}{c|}{LasHeR} & \multicolumn{2}{c|}{VisEvent} & \multicolumn{3}{c}{DepthTrack} \\
 &  &  &  &  & SR & PR & NPR & SR & PR & Pr & Re & F-score \\ \hline \hline
 &  &  & 92.13 & 56.44 & 0.479 & 0.590 & 0.555 & 0.412 & 0.515 & 0.536 & 0.522 & 0.529 \\
\checkmark &  &  & 93.96 & 57.44 & 0.567 & 0.709 & 0.671 & 0.595 & 0.761 & 0.583 & 0.566 & 0.571 \\
& \checkmark & &95.22 &58.06 & 0.523 &0.653 & 0.618 &0.578 &0.740 &0.572 &0.577& 0.569  \\
&  & \checkmark &92.45 & 56.57 &0.490 &0.601 &0.564&0.416 &0.519 &0.545 &0.555&0.560 \\
\checkmark & \checkmark & & 95.52 & 58.24 & 0.570 & 0.710 & 0.675 & 0.603 & 0.767 & 0.592 & 0.572 & 0.582 \\
&\checkmark & \checkmark &95.55 &58.14 &0.536 &0.665 &0.629 &0.580 &0.744 &0.581 &0.583  &0.579  \\
\checkmark&& \checkmark &92.74 & 56.75 &0.570 &0.712 &0.678 &0.597 &0.764 &0.593 &0.584 &0.587  \\
\checkmark & \checkmark & \checkmark & 95.85 & 58.32 & {\textbf{0.578}} & {\textbf{0.718}} & {\textbf{0.683}} & {\textbf{0.605}} &{\textbf{0.770}} & {\textbf{0.603}} & {\textbf{0.597}} & {\textbf{0.600}} \\ \hline
\end{tabular}
}
\end{table*}

\begin{table}[]
\renewcommand{\arraystretch}{1.5}
\centering
\caption{A comparison of different types of models on LasHeR test set. "Dim" denotes the dimension of the adapter modules. The value before the "\&" symbol typically represents the dimension of the hidden layer in HA and CEA, while the value after the "\&" symbol indicates the internal hidden layer dimension of MDA.
}\vspace{1mm}
\label{tab8}
\footnotesize
\scalebox{0.9}
{
\begin{tabular}{c|cc|c|cc|c}
\hline
Method & Params & Flops & Dim & SR & PR & FPS \\ \hline\hline
FFT & 92.13M & 56.44G & 768& 0.479 & 0.590 & 107.10 \\ 
ViPT & 92.97M & 21.8G &8 & 0.525 & 0.651 &24.78
\\ 
BAT & 92.45M&56.68G&8&0.563&0.702 
&20.22\\ 
SDSTrack &102.179M&108.34G & 8&0.531 &0.665&13.25 \\ 
{GMMT} & {962.2M} & {146.5G} &{768} & {0.566} &{0.707} &{17.40} \\
\hline
Our-tiny & 93.64M & 57.45G & 8\& 8 & 0.567 & 0.709 & 16.95\\
Our-base & {95.51M} & {58.24G} &8 \& 192  &{\textbf{{0.578}}}  &{\textbf{{0.718}}} &16.40 \\
Our-large & {102.1M} & {62.45G} & 192 \& 192 & 0.571 & 0.713 & 15.95 \\ \hline
\end{tabular}}
\end{table}

\begin{figure}[t]
\centering
\includegraphics[width=0.3\textwidth]{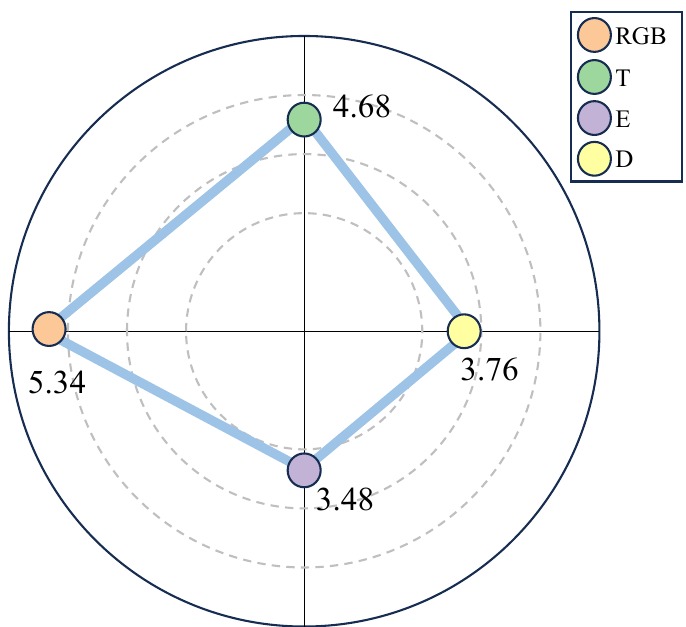} 
\caption{Exploration analysis of the informativeness of RGB, thermal, event, and depth modalities using single-modality information entropy. Single-modality information entropy refers to the entropy value calculated for each individual modality.}
\label{fig31}
\end{figure}

\textbf{GTOT.} 
GTOT \cite{GTOT} dataset is designed to assess the robustness of RGB-T tracking methodologies, consisting of 50 diverse sequences. As illustrated in Table \ref{tab2}, our approach demonstrates significant performance improvements relative to existing trackers. 
{The GMMT tracker with its fully fine-tuned network, achieves an impressive SR of 0.785 and PR of 0.936 on GTOT benchmark, surpassing the accuracy of PATrack.}
However, the adoption of generative techniques and a full fine-tuning mechanism significantly increases the network's parameters and computational cost, reaching 962.2M parameters and 146.5G FLOPs, as shown in Table \ref{tab8}. 
{In contrast, our proposed PATrack leverages an adapter-tuned mechanism, comprising 95.51M parameters and 58.24G FLOPs, which provides a significantly more efficient solution.}
When compared with the top-performing trackers, BAT\cite{bat} and SDSTrack \cite{sdstrack}, our proposed tracker presents the highest SR and PR by 2.0\%, 2.3\%, 1.6\% and 3.8\%.

\textbf{DepthTrack.} DepthTrack \cite{det} dataset is a comprehensive RGB-D tracking benchmark including 150 sequences for training and 50 sequences for testing.
{Although our results on DepthTrack dataset do not surpass SDSTrack, our understanding is that the depth data in DepthTrack dataset is sparse, unlike the RGB-T tracking tasks where thermal can provide richer target information, thus leading to a significant improvement in the metrics.}

{We also use information entropy to measure the average uncertainty within a probability distribution \cite{harte2014maximum,wang2024kcdnet}, and it could serve as a {valuable} indicator of redundancy inherent in different modalities. {To assess the relative informativeness of various modalities, we calculate the single-modality entropy for RGB, thermal, depth, and event data across  LasHeR, DepthTrack, and VisEvent benchmarks, as illustrated in Fig. \ref{fig31}.} RGB images yield the highest entropy score (5.34), reflecting their rich information content and structural complexity. 
Thermal images follow with an entropy of 4.68. In contrast, depth and event data display comparatively lower entropy values—3.76 and 3.48, respectively—indicating a higher degree of sparsity and reduced information density. This entropy-based comparison reinforces the observation that depth and event modalities inherently carry less information than thermal or RGB modalities.}

{Depth features exhibit spatial sparsity \cite{depthsparce}, with invalid areas (such as transparent surfaces) typically forming continuous blocks. 
This characteristic aligns well with the local masking strategy employed in the Complementary Masked Patch Distillation proposed by SDSTrack. 
As shown in \textit{Table 5} of the SDSTrack paper, the inclusion of Self-Distillation Learning (SD) notably enhances the model’s adaptability to depth data, resulting in a +1\% improvement in F-score. 
In contrast, the benefit for event data is smaller, with the SR increasing by only +0.2\%. 
Notably, SDSTrack only surpasses our method on depth data after incorporating SD, which constrains both clean and masked features—highlighting its effectiveness in addressing challenges specific to depth-based tracking. }

\begin{figure*}[t]
\centering
\includegraphics[width=1\linewidth]{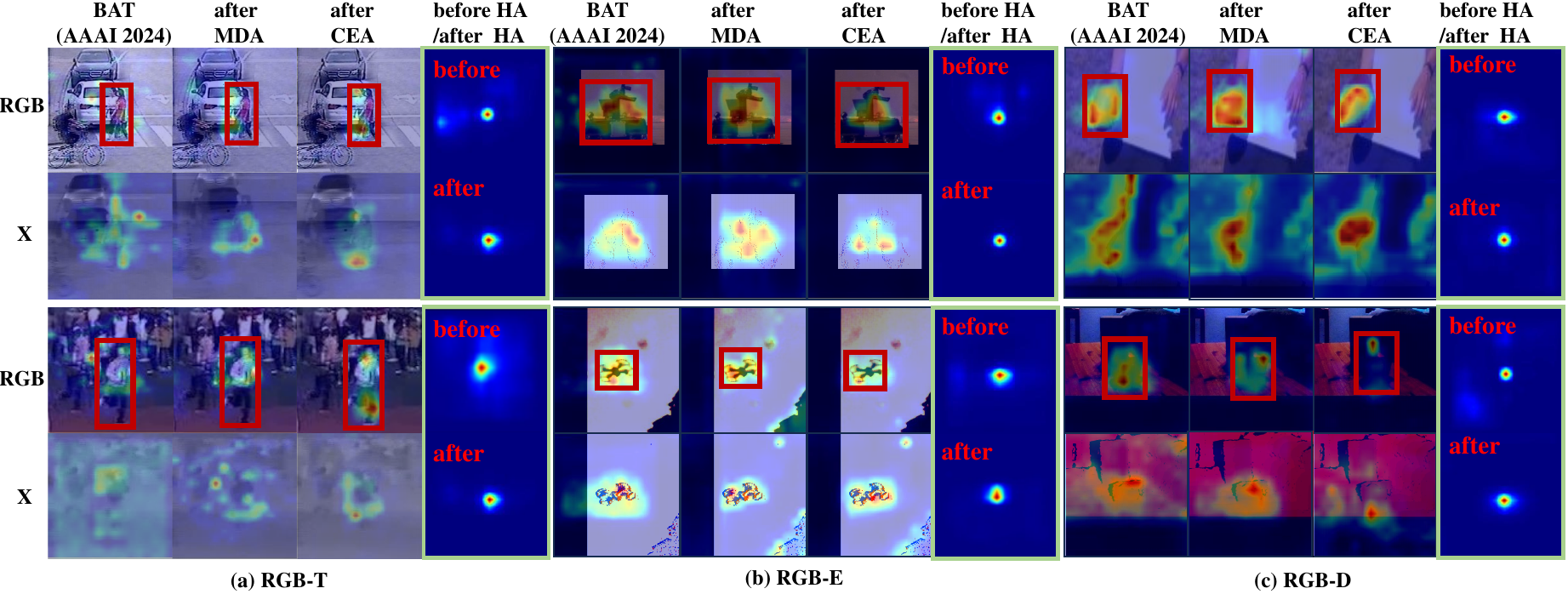}
\caption{Illustration of feature map comparison between BAT and our fusion mechanism and score maps comparison between before Head Adaptation and after Head Adaptation. 'after MDA' and ‘after CEA' represent the feature outputs of our proposed {Modality Dependent Adapter} and Cross-modality Entangled Adapter, respectively. (a) constitutes the visualization from LasHeR dataset (RGB-T). (b) corresponds to the visualization from VisEvent dataset (RGB-E), and (c) represents visualization from DepthTrack dataset (RGB-D). In every visualization group of \emph{before HA/after HA}, the top row shows the feature visualization without going through Head Adaptation, and the bottom row presents the visualization after going through Head Adaptation.
}
\vspace{-3mm}
\label{fig:1.1}
\end{figure*}

\textbf{VisEvent}. VisEvent \cite{visevent} dataset is the largest dataset available for RGB-E tracking, comprising 500 pairs of videos for training and 320 pairs for testing. 
We evaluate our proposed PATrack against current RGB-E tracking trackers. As shown in Table \ref{tab1}, our proposed tracker also surpasses previous EventVOT and CRSOT with improvements of 23.2\% and 8\% in SR and 22.4\% and 2.9\% in PR.
PATrack achieves great results, with the SR of 60.5 \% and PR of 77.0 \%.

\textbf{Attribute-Based Performance on LasHeR.}
{Our approach is evaluated across a diverse range of scenarios, with its performance analyzed on various attributes of LasHeR \cite{lasher} dataset. }
These attributes include No Occlusion (NO), Partial Occlusion (PO), Total Occlusion (TO), Hyaline Occlusion (HO), Motion Blur (MB), Low Illumination (LI), High Illumination (HI), Abrupt Illumination Variation (AIV), Low Resolution (LR), Deformation (DEF), Background Clutter (BC), Similar Appearance (SA), Camera Moving (CM), Thermal Crossover (TC), Frame Lost (FL), Out-of-View (OV), Fast Motion (FM), Scale Variation (SV), and Aspect Ratio Change (ARC).
As presented in Fig. \ref{fig3}, our approach outperforms existing trackers in most attributes for Precision Rate (PR).
In particular, in commonly challenging scenarios in SOT, such as HO, DEF, SV, MB, and CM, where the target undergoes severe deformation or is even temporarily invisible, our method can also achieve superior accuracy. 
Furthermore, whether the degradation occurs in RGB images, such as Low LI, HI, and AIV, or in TIR images, including TC and HO, our method excels.
This is attributed to the enhanced modality complementary information provided by the MDA module, the shared modality information integrated through the CEA module, and the HA module's effective adaptation to downstream multi-modal tasks. By enabling a more comprehensive interaction between RGB and TIR features, our method consistently delivers improved performance.

\begin{table}[]
\centering
\caption{The ablation experiment of MDA components on LasHeR dataset.}
\label{MDA}
\scalebox{1.2}{
\begin{tabular}{ccc|cc}
\hline
Avgpool & Maxpool & DWConv & SR & PR \\ \hline \hline
\multicolumn{1}{l}{} & \multicolumn{1}{l}{} & \multicolumn{1}{l|}{} & \multicolumn{1}{l}{0.536} & \multicolumn{1}{l}{0.665} \\ 
\checkmark &  &  & 0.542 & 0.676 \\
 & \checkmark &  & 0.535 & 0.667 \\
 &  & \checkmark &0.532 & 0.669  \\
\checkmark & \checkmark &  & 0.568 & 0.706 \\
 & \checkmark & \checkmark & 0.560 &0.673 \\
\checkmark  &  &\checkmark  & 0.562 & 0.691\\
\checkmark & \checkmark & \checkmark & \textbf{0.578} & \textbf{0.718} \\ \hline
\end{tabular}
}
\end{table}

\begin{table}[]
\centering
\caption{The ablation experiment of CEA components on LasHeR dataset.}
\label{CEA}
\scalebox{1.3}
{
\begin{tabular}{c|cc}
\hline
Method & SR & PR \\ \hline \hline
w/o CEA & 0.570 &0.712  \\
\multicolumn{1}{l|}{w/o Fusuion-guided} & \multicolumn{1}{l}{0.572} & \multicolumn{1}{l}{0.713} \\
w/o Conv & 0.574 & 0.714 \\
w/o Skip Connect & 0.575 & 0.714 \\
w Full CEA & \textbf{0.578} & \textbf{0.718} \\ \hline
\end{tabular}
}
\end{table}

\subsection{Ablation Study}
\textbf{Component analysis of PATrack.} {The results from ablation experiments on LasHeR, VisEvent, and DepthTrack benchmarks are detailed in Table~\ref{tab7}, highlighting the influence of our method components.}
We analyze to validate the effectiveness of the proposed MDA, CEA, and HA modules. Our baseline model employs a dual-branch structure, OSTrack, with feature fusion implemented in the last layer of the Transformer block. 
As shown in Table \ref{tab7}, the incorporation of MDA results in significant improvements, with a 4.2\% increase in the F-score on DepthTrack compared to the baseline. 
{Furthermore, the SR and NPR on LasHeR increase by 8.8\% and 11.6\%, respectively, and the SR and PR on VisEvent improve by 28.3\% and 24.6\%, respectively.}
Further combination with CEA enhances effectiveness, notably increasing the F-score on DepthTrack by 1.1\%. The addition of HA also improves performance, with a 0.8\% boost in SR on LasHeR and a 1.8\% F-score increase on DepthTrack. 
Additionally, the SR on the VisEvent experiences a slight increase of 0.2\%.
Due to the inherent sparsity of event data, {which differs from thermal data that encompasses a richer set of information, the improvement in HA accuracy} is relatively minor.

\textbf{Different hidden layer dimensions in adapters.} 
We explore the effectiveness of different hidden layer dimensions for three types of adapters. 
We select various combinations of 8 and 192 as the dimensions for our adapter. 
The choice of 8 is based on its common use as a dimension in adapters, facilitating a fair comparison with benchmark methods such as ViPT, BAT, and SDSTrack. 
The selection of 192 is due to the fact that attention mechanisms typically require a sufficient number of parameters {to calculate the weights between different input elements accurately.}
A larger dimension provides more parameters, enabling a more refined distribution of attention. 
{From Table \ref{tab8}, it can be observed that our method achieves more accurate localization and tracking of the target with only a slight increase in the number of parameters and computational load.}

\textbf{Component analysis of the MDA module.}
Considering the effectiveness and efficiency of high- and low-frequency components, we analyze the proposed MDA module in Table \ref{MDA}. As shown, introducing operations like max pooling, depthwise separable convolution, or average pooling individually enhances network performance compared to configurations without the MDA module. Notably, adding average pooling yields the most significant improvement, as Transformers inherently retain more {low-frequency information.} Combining Avgpool and Maxpool results in better performance compared to combinations of Maxpool with DWConv or Avgpool with DWConv. This is likely because the latter configurations either prioritize high-frequency extraction while neglecting low-frequency enhancement or fail to effectively capture critical high-frequency information when relying solely on Maxpool. When Maxpool, Avgpool, and DWConv are combined, the evaluation metrics for SR and PR reach higher values of 0.578 and 0.718, respectively, validating that DWConv enhances high-frequency feature extraction in conjunction with Maxpool.

\begin{figure*}[t]
\centering
\includegraphics[width=1.8\columnwidth]{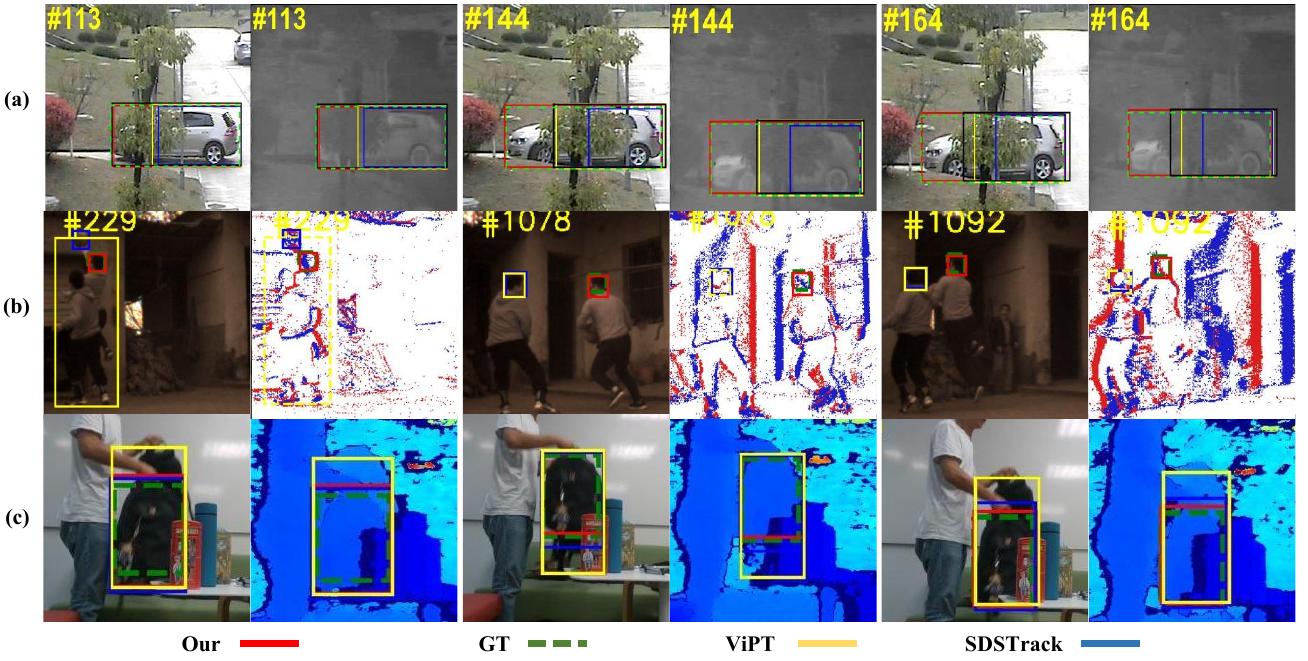} 
\caption{Tracking results of RGB-T, RGB-E, and RGB-D. (a) Represents tracking results of RGB-T tracking derived from the LasHeR dataset. (b) Represents tracking results of RGB-E tracking derived from the VisEvent dataset. (c) Represents tracking results of RGB-D tracking derived from the DepthTrack dataset.}
\label{fig5}
\end{figure*}

\begin{table}[]
\caption{Results of different CEA adapter layers in the LasHeR. 
'Even' and 'uneven' respectively represent the positions where ECA is evenly and unevenly placed.
'2', '4', and '6' represent replacing the MDA module with the CEA module every 2, 4, and 6 layers, respectively.
}
\vspace{1mm}
\label{tab9}
\renewcommand{\arraystretch}{1.2}
\centering
\scalebox{1.2}
{
\begin{tabular}{cc|ccc}
\hline
\multicolumn{2}{c|}{{Layers}} & {SR} & {PR} & {NPR} \\ \hline \hline
\multicolumn{2}{c|}{w/o CEA} & \multicolumn{1}{l}{0.574} & \multicolumn{1}{l}{0.715} & \multicolumn{1}{l}{0.679} \\ \hline
\multicolumn{1}{c|}{\multirow{3}{*}{even}} & 2 & 0.573 & 0.715 & 0.678 \\
\multicolumn{1}{c|}{} & 4 & 0.562 & 0.697 & 0.663 \\
\multicolumn{1}{c|}{} & 6 & 0.559 & 0.696 & 0.661 \\ \hline
\multicolumn{1}{c|}{\multirow{7}{*}{\begin{tabular}[c]{@{}c@{}}un-\\ even\end{tabular}}} & 4 & 0.567 & 0.699 & 0.664 \\
\multicolumn{1}{c|}{} & 7 & 0.562 & 0.697 & 0.663 \\
\multicolumn{1}{c|}{} & 10 & 0.560 & 0.671 & 0.662 \\
\multicolumn{1}{c|}{} & 4,7 & 0.567 & 0.706 & 0.669 \\
\multicolumn{1}{c|}{} & 7,10 & 0.574 & 0.721 & 0.658 \\
\multicolumn{1}{c|}{} & {4,10} &0.566  &0.690  & 0.668\\
\multicolumn{1}{c|}{} & 4,7,10 & {\textbf{0.578}} & {\textbf{0.718}} & {\textbf{0.683}} \\ \hline
\end{tabular}
}
\end{table}

\textbf{Component analysis of the CEA module.}
To evaluate the effectiveness of the CEA module, we examine its components as detailed in Table \ref{CEA}.


\emph{w/o Fusion-guided} represents the absence of multi-modal-guided information, where the cross-attention mechanism in the original CEA module transitions to a self-attention operation. Compared with the baseline lacking the CEA module (\emph{w/o CEA}), this structure demonstrates improved performance, indicating its ability to capture a certain degree of multi-modal information. 
However, compared with our proposed full CEA module, its PR and SR decrease by 0.6\% and 0.5\%, respectively. This performance degradation is attributed to the lack of cross-modal correlation information, which hinders the enhancement of discriminative capability and limits overall performance.

\emph{w/o Conv} eliminates convolution layers that process RGB and X modalities to extract Q, K, and V data. Nonetheless, compared with \emph{w Full CEA}, its SR and PR decline by 0.4\% and 0.4\%, respectively. 
Since the convolution layer can enhance the model representational ability, it still achieves higher accuracy than \emph{w/o CEA}. 

{\emph{w/o Skip Connection} omits the operation of adding original modality data back into the cross-attention calculation. }
This omission leads to inferior performance compared with the \emph{w Full CEA}, as the skip connection further enhances shared modality information and improves cross-modal discriminative capability.

\textbf{Impact of inserted layers.}
{Considering the effectiveness and efficiency of the network, we investigate the even and uneven distribution of proposed modules in Table \ref{tab9}. }
For an even distribution, placing the CEA module every two layers rather than every four or six layers results in a higher number of modules and a stronger modality correlation. 
This adjustment significantly enhances performance, as it allows for more frequent interactions and deeper integration of complementary multi-modal information across the network. However, this improvement does not surpass the performance of the network without the addition of the CEA module.

{As for uneven distribution, previous work \cite{layerselect} identifies the homogeneous nature of Transformer architectures, and we strategically place our CEA modules at representative layers (4th, 7th, and 10th).
This placement enables effective hierarchical feature extraction across distinct network depths including shallow, middle, and deep levels.}
{Then, this layer selection follows existing practices in multi-modal tracking research (e.g., TBSI), where these specific layers have proven critical operations like candidate elimination and cross-modal interaction. 
Additionally, when CEA modules are unevenly distributed, placed at layers 4, 7, and 10, the performance exceeds that of the network without CEA in terms of SR, PR, and NPR by 0.4\%, 0.3\%, and 0.4\%, respectively. It hints that the model engages in a more sophisticated and extensive joint learning process, effectively integrating both intra-modal and inter-modal information.
}
Analysis from {Table} \ref{tab9} indicates that performance declines when only one of the 4th, 7th, or 10th layers replaces its MDA module with a CEA module.
However, equipping two layers, particularly the 7th and 10th, with CEA modules yields substantial performance improvements.
When CEA is inserted into all three layers—the 4th, 7th, and 10th, the PR, SR, and NPR achieve their highest values.
{In summary, the 4/7/10 configuration remains superior by maintaining an optimal trade-off between model performance and computational efficiency. }

\subsection{Visualization}

\textbf{Effectiveness of Progressive Adaptation.} 
{We conduct a qualitative analysis of the proposed PATrack components for various tracking tasks including RGB-T, RGB-D, and RGB-E.}
As illustrated in Fig. \ref{fig:1.1}, the MDA module effectively enhances specific modality features for both RGB and X modalities, improving the attention distribution toward the target. Despite the differences in target distributions between RGB and X modalities, the addition of the CEA module, which leverages modality correlations, enables the utilization of shared regions of interest. This integration facilitates effective cross-modal fusion, further enhancing tracking performance.
Regarding the visualization of features processed through the HA module, this module effectively filters out background information, even when the data in RGB-E and RGB-D modalities are inherently sparse.
This further indicates that the RGB pre-trained prediction head architecture is not well-suited for downstream multi-modal tasks, and our proposed specific adapter effectively enhances the capability to adapt to these downstream tasks.

\textbf{Visualization of tracking results.}
We compare the tracking performance of our method with ViPT and SDSTrack in Fig. \ref{fig5}.
It can be observed that in some challenging sequences, such as (a) sequence with occlusions, (b) sequence where the target appears off-screen multiple times and moves rapidly, and (c) sequence with background interference at different depths, our method can achieve more accurate performance. 
{This indicates that our method could fully utilize multi-modal complementary information and modality-related information.}

\begin{figure}[t]
\centering
\includegraphics[width=0.46\textwidth]{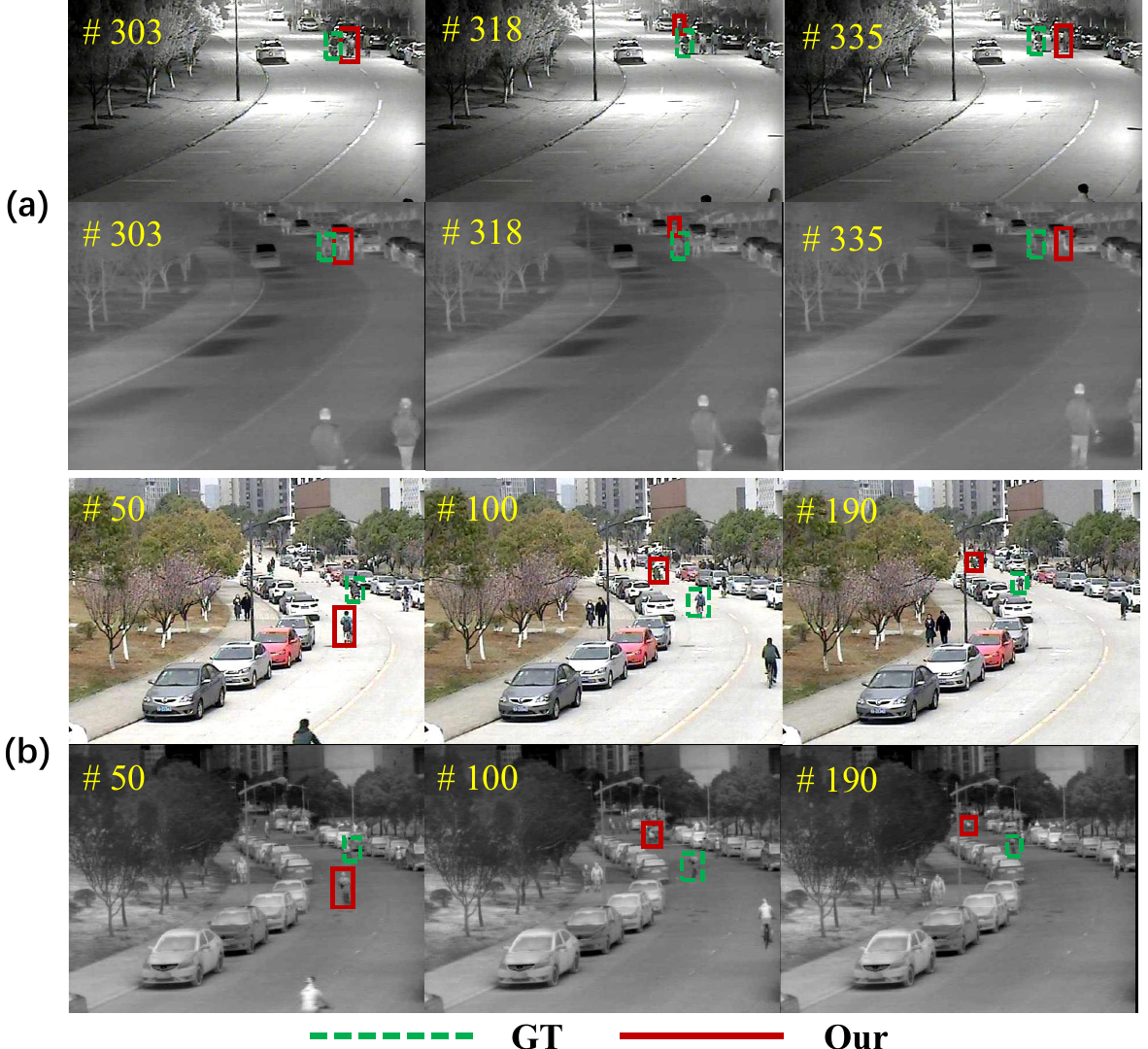} 
\caption{{
{Illustration of failure cases on RGBT234 dataset.}
(a) shows the tracking results from the "bike" sequence. (b) corresponds to the "boundaryandfast" sequence. For each group, the top row presents the RGB images, and the bottom row shows the corresponding thermal images.}}
\label{fig6}
\end{figure}

{\subsection{Limitation Analysis and Future Expectation}}
{\textbf{Analysis of Failure Cases.} Fig. \ref{fig6} (a) illustrates a tracking failure case from the "bike" sequence. When the target bike appears in the upper right corner of the frame, the combination of low-light condition and the presence of visually similar objects in the surroundings leads to a degradation in the tracker’s robustness in subsequent frames. 
Fig. \ref{fig6}(b) presents another failure case from the “boundaryandfast” sequence, where the target is a person riding a motorcycle. As the target moves into a cluttered background, our tracker mistakenly switches to another person riding a different bike.
These failures are partially attributed to the use of an offline tracker, which relies on the first frame as the template. 
This fixed-template approach is vulnerable to failure when the target or scene undergoes significant changes, especially in long video sequences. 
To mitigate this limitation, we plan to incorporate a template update mechanism that updates the template every 25 frames, effectively converting the tracker into an online framework and improving its adaptability to dynamic environments.}

{\textbf{Limitation of tracking efficiency.} To enhance the efficiency of our PATrack, we plan to propose a solution that combines module-level structured pruning and dynamic sparsification. First, we employ a head pruning strategy in the CEA module based on gradient sensitivity analysis. 
Specifically, the number of attention heads is reduced from 8 to 4 by evaluating the contribution of each head to tracking precision rate and retaining the four most impactful heads. 
This approach significantly reduces computational overhead while minimizing performance degradation. Second, we further plan to introduce a dynamic sparsification mechanism. 
{A lightweight CNN-based modality quality evaluator, comprising two convolutional layers and one MLP layer, can be designed to dynamically output clarity scores (ranging from 0 to 1) for both RGB and thermal modalities.} 
Based on these scores, we implement adaptive computation rules:
(1)	If the RGB quality score exceeds 0.7 and the thermal imaging score is below 0.3, the CEA module computation is skipped, retaining only the MDA module computation for the RGB branch.
(2)	If the thermal modality score exceeds 0.6, all CEA layers are enforced to leverage the thermal advantage.
(3)	In other cases, all adapter modules execute as normal.
This solution balances efficiency and accuracy by leveraging pruning and dynamic computation, offering a significant improvement in computational efficiency while maintaining robust tracking performance.
}

{\textbf{Future Expectation.} Although our method demonstrates satisfactory performance and achieves the unified network architecture across different multi-modal tracking tasks, it still has the limitation of requiring multiple training processes for each individual task and the achievement of a one-time training process with adaptive recognition of input data. 
Additionally, the significant differences in the richness and characteristics of depth, event, and thermal data further complicate the tracking process. 
Our current method does not incorporate modality customization, which limits its flexibility and applicability across various tasks. 
To address these challenges, our future work will focus on developing a unified tracking model that includes modality-specific customization to enhance performance.
This will enable the seamless integration of different modalities, allowing the algorithm to better handle variations in tasks and potentially improve tracking accuracy across diverse environments.

\section{Conclusion}
This paper proposes a novel approach, Progressively Learning Adaptation for Multi-modal Tracking (PATrack), to enhance the adaptability of pre-trained networks for downstream tasks.
Our proposed PATrack demonstrates outstanding performance against other trackers across challenging datasets, including LasHeR, GTOT, RGBT234, VisEvent, and DepthTrack benchmarks, substantiating its effectiveness through a series of RGB SOT downstream tasks, including RGB+Thermal, RGB+Event, and RGB+Depth tracking.
We attribute this enhanced performance to the complementary learning of inter-modal and intra-modal information from multi-modal feature maps.
Besides, the prediction head architecture is optimized to effectively bridge the gap between the pre-trained network and the downstream task of multi-modal tracking.




\bibliographystyle{IEEEtran}
\bibliography{ref}

\begin{IEEEbiography}[{\includegraphics[width=0.9 in, keepaspectratio]{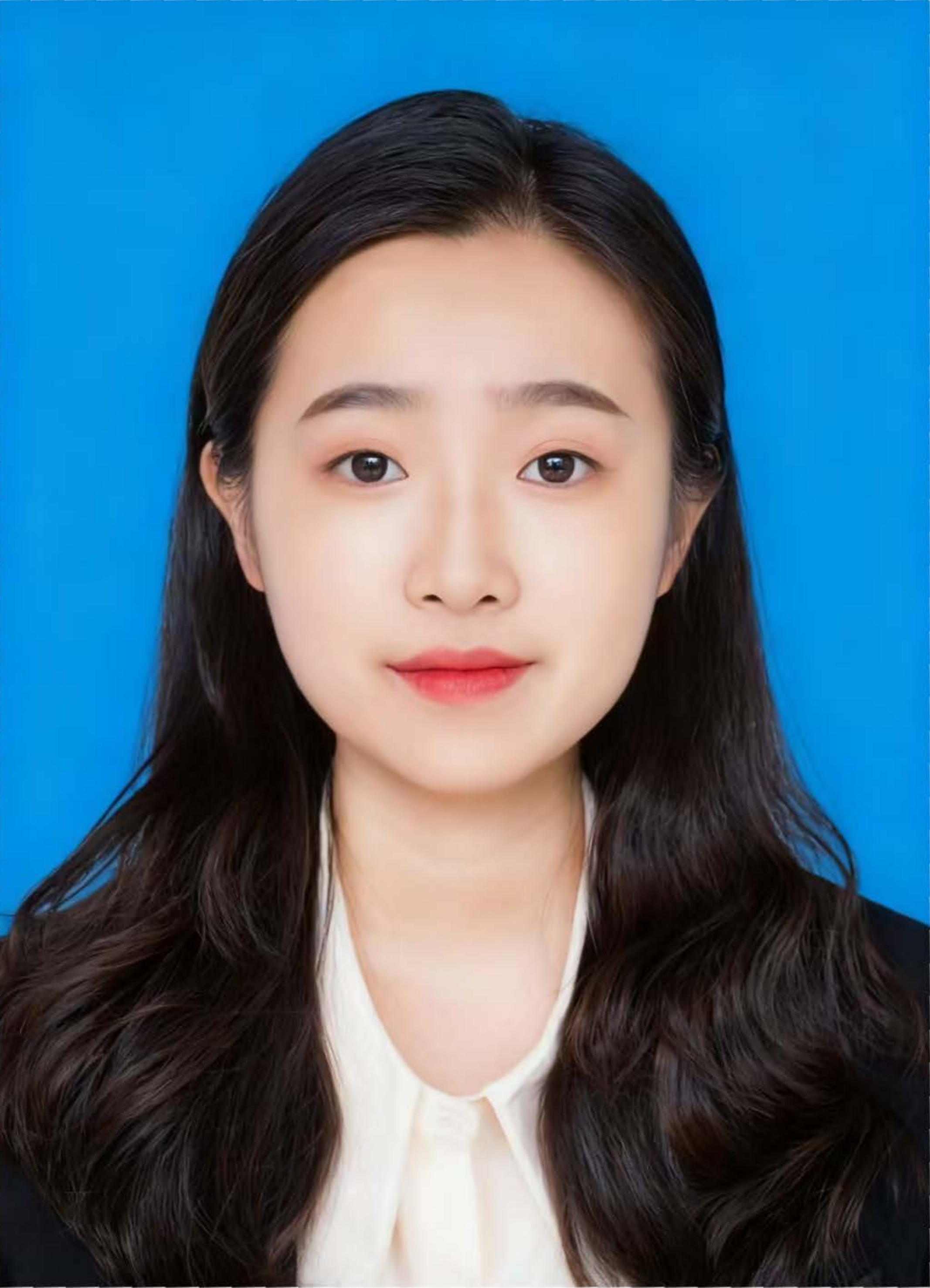}}]{He Wang}
is now a Ph.D. student with the School of Artificial Intelligence and Computer Science, Jiangnan University. Her research interests include multi-modal object tracking and deep learning.
\end{IEEEbiography}

\begin{IEEEbiography}[{\includegraphics[width=0.9 in, keepaspectratio]{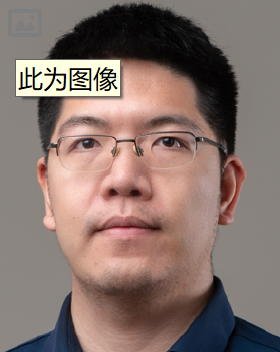}}]{Tianyang Xu}
 (Member, IEEE) received the B.Sc. degree in electronic science and engineering from Nanjing University, Nanjing, China, in 2011, and the Ph.D. degree from the School of Artificial Intelligence and Computer Science, Jiangnan University, Wuxi, China, in 2019. He is currently an Associate Professor with the School of Artificial Intelligence and Computer Science, Jiangnan University. His research interests include visual tracking and deep learning.
\end{IEEEbiography}


\begin{IEEEbiography}[{\includegraphics[width=0.9 in, keepaspectratio]{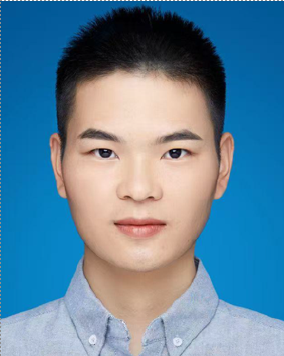}}]{Zhangyong Tang}
is now a Ph.D. student with the School of Internet of Things Engineering, Jiangnan University. His research interests include multi-modal object tracking and deep learning.
\end{IEEEbiography}


\begin{IEEEbiography}[{\includegraphics[width=0.9 in, keepaspectratio]{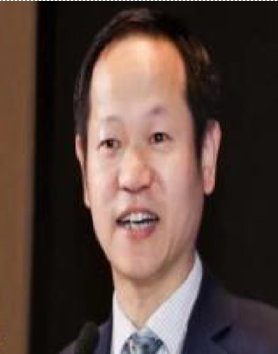}}]{Xiao-Jun Wu}
received the B.Sc. degree in mathematics from Nanjing Normal University, Nanjing, China, in 1991, and the M.S. and Ph.D. degrees in pattern recognition and intelligent systems from Nanjing University of Science and Technology, Nanjing, in 1996 and 2002, respectively. He is currently a Professor in artificial intelligence and pattern recognition with Jiangnan University, Wuxi, China. His research interests include pattern recognition, computer vision, fuzzy systems, neural networks, and intelligent systems. He is currently a fellow of IAPR and AAIA. He has won several domestic
and international awards because of his research achievements. He served as an associate editor for several international journals.
\end{IEEEbiography}


\begin{IEEEbiography}[{\includegraphics[width=0.9 in, keepaspectratio]{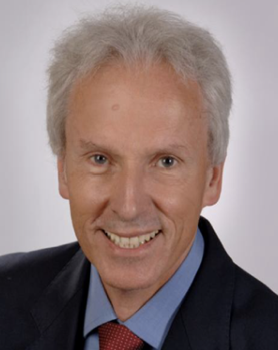}}]{Josef Kittler}
 received the B.A., Ph.D., and D.Sc. degrees from the University of Cambridge, in 1971, 1974, and 1991, respectively. He is a distinguished Professor of Machine Intelligence at the Centre for Vision, Speech and Signal Processing, University of Surrey, Guildford, U.K. He conducts research in biometrics, video and image database retrieval, medical image analysis, and cognitive vision. He published the textbook Pattern Recognition: A Statistical Approach and about 1000 scientific papers. His publications have been cited by around 70,000 times.

He is series editor of Springer Lecture Notes on Computer Science. He currently serves on the Editorial Boards of Pattern Recognition Letters, Pattern Recognition and Artificial Intelligence, Pattern Analysis and Applications. He also served as a member of the Editorial Board of IEEE Transactions on Pattern Analysis and Machine Intelligence during 1982-1985. He served on the Governing Board of the International Association for Pattern Recognition (IAPR) as one of the two British representatives during 1982-2005, and the President of IAPR during 1994-1996.
\end{IEEEbiography}

\end{document}